\documentclass[acmtog, authorversion]{acmart}

\usepackage{booktabs}
\usepackage{epsfig}
\usepackage{graphicx}
\usepackage{subcaption}
\usepackage{amsfonts}
\usepackage{amsmath}
\usepackage{gensymb}
\usepackage{enumitem}
\usepackage{bm}
\usepackage{tikz-cd}

\usepackage{soul} \setstcolor{red}

\newcommand{\g}{\mathbf{\Gamma}}

\setcitestyle{square}

\setcopyright{rightsretained}
\acmJournal{TOG}
\acmYear{2022}\acmVolume{41}\acmNumber{4}\acmArticle{105}\acmMonth{7} \acmDOI{10.1145/3528223.3530166}
\begin{document}

% Title.
\title{DeltaConv: Anisotropic Operators for Geometric Deep Learning on Point Clouds}

% Authors.
\author{Ruben Wiersma}
\orcid{0000-0001-7900-7253}
\email{r.t.wiersma@tudelft.nl}
\affiliation{%
 \institution{Delft University of Technology}
 \country{The Netherlands}
 }
\author{Ahmad Nasikun}
\orcid{0000-0001-5311-4456}
\email{a.nasikun@tudelft.nl}
\affiliation{%
 \institution{Delft University of Technology}
 \country{The Netherlands}
 \institution{ and Universitas Gadjah Mada}
 \country{Indonesia}
 }
\author{Elmar Eisemann}
\orcid{0000-0003-4153-065X}
\email{e.eisemann@tudelft.nl}
\affiliation{%
 \institution{Delft University of Technology}
 \country{The Netherlands}
 }
\author{Klaus Hildebrandt}
\orcid{0000-0002-9196-3923}
\email{k.a.hildebrandt@tudelft.nl}
\affiliation{%
 \institution{Delft University of Technology}
 \country{The Netherlands}
 }
 
\renewcommand\shortauthors{R. Wiersma, A. Nasikun, E. Eisemann, \& K. Hildebrandt}

% abstract
\begin{abstract}
Learning from 3D point-cloud data has rapidly gained momentum, motivated by the success of deep learning on images and the increased availability of 3D~data. In this paper, we aim to construct anisotropic convolution layers that work directly on the surface derived from a point cloud. This is challenging because of the lack of a global coordinate system for tangential directions on surfaces. We introduce DeltaConv, a convolution layer that combines geometric operators from vector calculus to enable the construction of anisotropic filters on point clouds. Because these operators are defined on scalar- and vector-fields, we separate the network into a scalar- and a vector-stream, which are connected by the operators. The vector stream enables the network to explicitly represent, evaluate, and process directional information. Our convolutions are robust and simple to implement and match or improve on state-of-the-art approaches on several benchmarks, while also speeding up training and inference.
\end{abstract}

%CCS
\begin{CCSXML}
<ccs2012>
<concept>
<concept_id>10010147.10010257.10010293.10010294</concept_id>
<concept_desc>Computing methodologies~Neural networks</concept_desc>
<concept_significance>500</concept_significance>
</concept>
<concept>
<concept_id>10010147.10010371.10010396.10010402</concept_id>
<concept_desc>Computing methodologies~Shape analysis</concept_desc>
<concept_significance>500</concept_significance>
</concept>
</ccs2012>
\end{CCSXML}

\ccsdesc[500]{Computing methodologies~Neural networks}
\ccsdesc[500]{Computing methodologies~Shape analysis}

%keywords
\keywords{Point Clouds, Point Cloud Learning, Point Cloud Processing, Geometric Deep Learning, Graph CNN}

\maketitle

\section{Introduction}
The success of convolutional neural networks (CNNs) on images and the increasing availability of point-cloud data motivate generalizing CNNs from images to 3D point clouds \cite{guo2020survey, s19194188, Bronstein2017}. One way to achieve this is to design convolutions that operate directly on the surface. Such \textit{intrinsic} convolutions reduce the kernel space to tangent spaces, which are two-dimensional on surfaces. Compared to extrinsic convolutions, intrinsic convolutions can be more efficient and the search space for kernels is reduced, they naturally ignore empty space, and they are robust to rigid- and non-rigid deformations \cite{Boscaini2016}. Examples of intrinsic convolutions on point clouds are GCN~\cite{kipf2017}, PointNet++~\cite{Qi2017b}, EdgeConv~\cite{Wang2019}, and DiffusionNet~\cite{Sharp2020DiffusionIA}. 

\begin{figure}
    \centering
    \includegraphics[width=0.85\columnwidth]{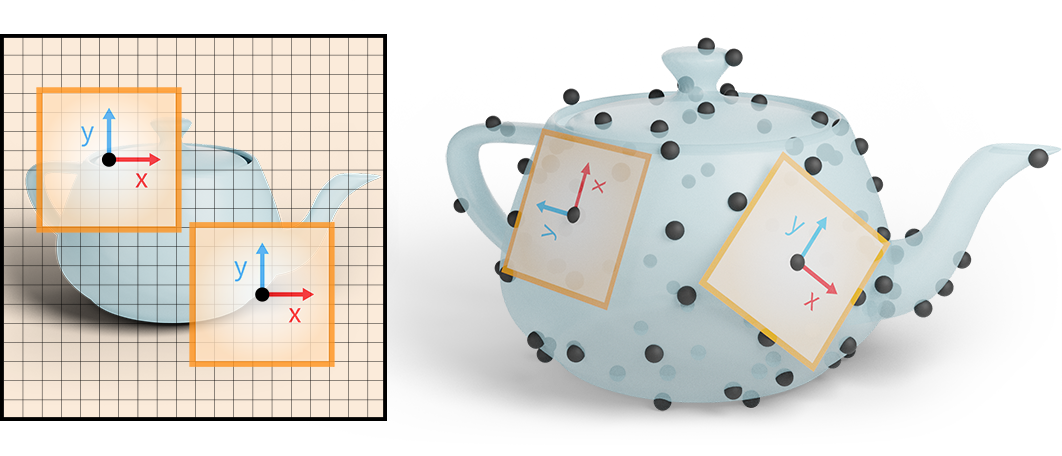}
    \caption{Images have a global coordinate system (left). Point clouds do not (right), complicating the design of anisotropic convolutions.}
    \label{fig:intro}
\end{figure}

Our focus is on constructing intrinsic convolutions which are anisotropic or direction-dependent. This is difficult because of the fundamental challenge that non-linear manifolds lack a global coordinate system.
As an illustration of the problem, consider a CNN on images (\autoref{fig:intro}, left).
Because an image has a globally consistent up-direction, the network can build anisotropic filters that activate the same way across the image.
For example, one filter can test for vertical edges and the other for horizontal edges. No matter where the edges are in the image, the filter response is consistent. In subsequent layers, the output of these filters can be combined, e.g., to find a corner.
Because we do not have a global coordinate system on surfaces (\autoref{fig:intro}, right), one cannot build and use anisotropic filters in the same way as on images. This limits current intrinsic convolutions on point clouds. For example, GCNs filters are isotropic. PointNet++ uses maximum aggregation and adds relative point positions, but still applies the same weight matrix to each neighboring point.

We introduce a new way to construct anisotropic convolution layers for geometric CNNs. Our convolutions are described in terms of geometric operators instead of kernels. The operator-based perspective is familiar from GCN, which uses the Laplacian on graphs. While the Laplacian is a natural fit for intrinsic learning on surfaces, it is isotropic. A classical way of creating anisotropic operators is to write the Laplacian as the divergence of the gradient and apply a linear or non-linear operation on the intermediate vector field~\cite{weickert1998anisotropic}. We build on this idea by constructing learnable anisotropic operators from elemental geometric operators: the gradient, co-gradient, divergence, curl, Laplacian, and Hodge-Laplacian. These operators are defined on spaces of scalar fields and tangential vector fields. Hence, our networks are split into two streams: one stream contains scalars and the other tangential vectors. The operators map along and between the two streams. The vector stream encodes feature activations and directions along the surface, allowing the network to test and relate directions in subsequent layers. Depending on the task, the network outputs scalars or vectors.
A property of a network constructed from these operators is that it is coordinate-independent: though bases of the tangent spaces of a point cloud need to be chosen, the weights learned by the network will be the same no matter what bases are chosen. 
Hence, we can realize direction-dependent convolutions despite the lack of global coordinate systems on surfaces and without the need of specially constructed tangent space bases. We name our convolutions \textit{DeltaConv}.

\begin{figure}
    \centering
    \includegraphics[width=0.95\columnwidth]{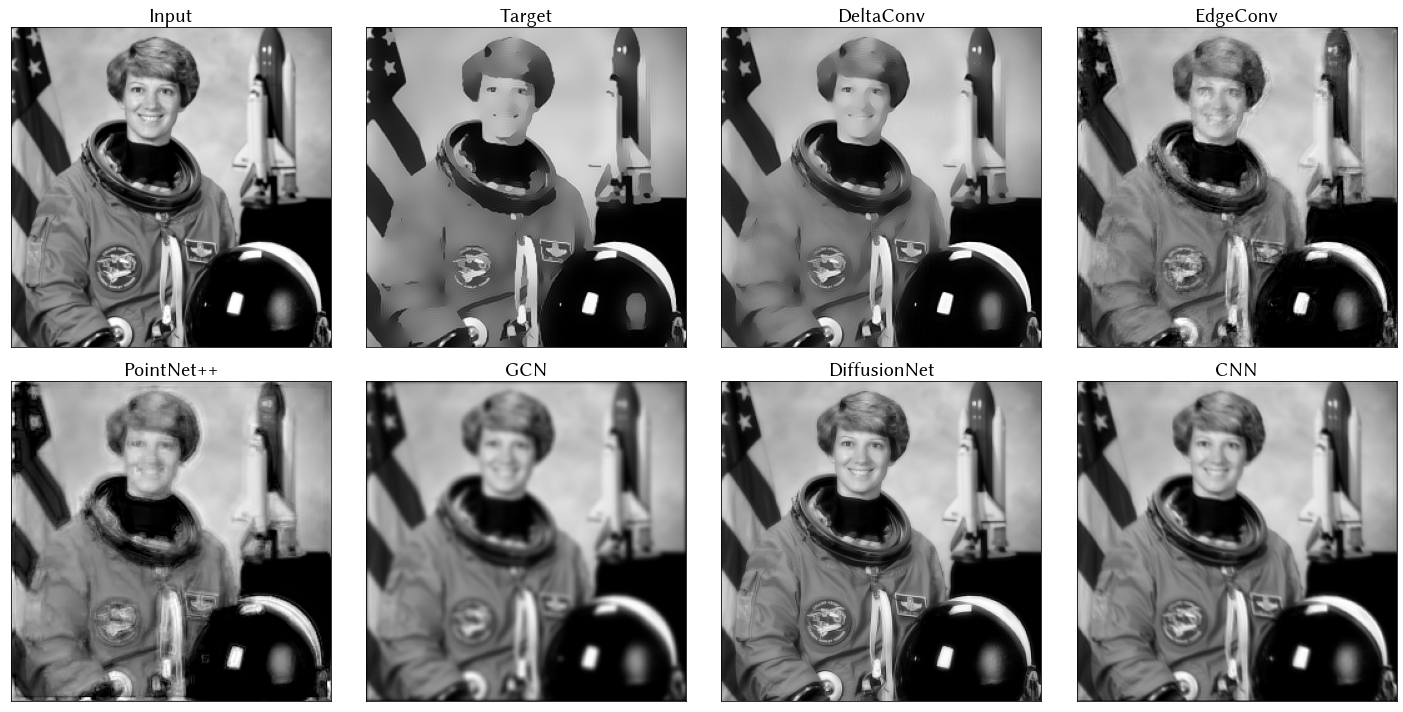}
    \caption{A ResNet with varying convolutions is overfitted to a target image created with twenty anisotropic diffusion steps. DeltaConv can reproduce the filter well, where other convolutions struggle. (Courtesy NASA)}
    \label{fig:peronamalik}
\end{figure}

To get an idea of the benefits of DeltaConv, consider the anisotropic image filter proposed by Perona and Malik \shortcite{peronamalik}. The Perona--Malik filter integrates an anisotropic diffusion equation in which the anisotropic operator combines the gradient, a non-linearity, and the divergence. 
DeltaConv has access to the building blocks needed to construct such an anisotropic operator and to perform explicit integration steps of the diffusion equation.
This is illustrated in \autoref{fig:peronamalik}. We trained a simple ResNet \cite{he2016deep} to match the result of twenty anisotropic diffusion steps on a sample image. While DeltaConv can reproduce the filter well, other intrinsic convolutions and regular image convolutions fail to capture the effect, producing overly smooth signals or artifacts instead.
Additional benefits of our approach are the following: by maintaining a stream of vector features throughout the network, our convolutions can relate directional information between different points on the surface. Together with the increased expressiveness of convolutions due to anisotropy, this results in increased accuracy over isotropic convolutions, as well as state-of-the-art approaches, as we show in our experiments. Also, each operator is implemented as a sparse matrix and the combination of operators is computed per point, which is simple and efficient. 

In our experiments, we demonstrate that a simple architecture with only a few DeltaConv blocks can match and, in some cases, outperform state-of-the-art results using more complex architectures. We achieve 93.8\% accuracy on ModelNet40, 84.7\% on the most difficult variant of ScanObjectNN, 86.9 mIoU on ShapeNet, and 99.6\% on SHREC11, a dataset of non-rigidly deformed shapes. Our ablation studies show that adding the vector stream can decrease the error by up to 25\% (from 90.4\% to 92.8\%) on ModelNet40 and up to 21\% for ShapeNet (from 81.1\ to 85.1 mIoU), while the use of per-point directional features speeds up inference by $1.5-2\times$ and the backward pass by $2.5-30\times$ compared to edge-based features.

Summarizing our main contributions:
\begin{itemize}
    \item We introduce a new construction of convolution layers for geometric CNNs that supports the construction of anisotropic filters.
    This is achieved by letting networks learn convolutions as compositions and linear combinations of geometric differential operators and point-wise non-linearities. Moreover, the networks maintain a stream of vector features in addition to the usual stream of scalar features and use the operators to communicate in and between the streams.
    \item We propose a network architecture that realizes our approach and adapt the differential operators to work effectively in our networks.
    \item We implement and evaluate the network for point clouds\footnote{The implementation is available at \url{https://github.com/rubenwiersma/deltaconv}.}
    and propose techniques to cope with undersampled regions, noise, and missing information prevalent in point cloud learning.
\end{itemize}

\section{Related work}
We focus our discussion of related work on the most relevant topics. Please refer to surveys on geometric deep learning~\cite{Bronstein2017, bronstein2021geometric} and point cloud learning~\cite{guo2020survey, s19194188} for a more comprehensive overview of this expanding field.
\paragraph*{Point cloud networks and anisotropy}
A common approach for learning on point-cloud data is to learn features for each point using a multi-layer perceptron (MLP), followed by local or global aggregation. Many methods also learn features on local point pairs before maximum aggregation.
Well-known examples are PointNet and its successor PointNet++~\cite{Qi2017a, Qi2017b}. Several follow-up works improve speed and accuracy, for example by adding more combinations of point-pair features~\cite{Zhao_2019_CVPR, Sun2019, Yang2019ModelingPC, Le_2020_CVPR, closerlook, drnet, gdanet, cganet}. Some of these point-wise MLPs explicitly encode anisotropy by splitting up the MLP for each 3D axis \cite{Lan2019ModelingLG, closerlook}. Concepts from transformers~\cite{vaswani2017attention} have also made their way to point clouds~\cite{pointtransformer, pointvoxeltransformer, lin2020onepoint}. These networks use self-attention to compute aggregation weights for (neighboring) points. Spatial information is incorporated by adding relative positions in 3D. Attention-based aggregation could be used in our approach as a replacement of maximum aggregation. The distance between points could serve as an intrinsic spatial encoding.

Pseudo-grid convolutions are a more direct translation of image convolutions to point clouds. Many of these are defined in 3D and thus support anisotropy in 3D coordinates. Several works learn a continuous kernel and apply it to local point-cloud regions~\cite{liu2019rscnn, Boulch2019ConvPoint, Densepoint2019, thomas2019KPConv, Wu_2019_CVPR, Hermosilla2018, Matan2018,Fey2018, paconv}. Others learn discrete kernels and map points in local regions to a discrete grid~\cite{Hua2018, lei2019octree, li2018pointcnn, minkowskinet, SparseConvNet}. We go into an orthogonal direction by building intrinsic convolutions, which operate in fewer dimensions and naturally generalize to (non-)rigidly deformed shapes.

Finally, graph-based approaches create a k-nearest neighbor- or radius-graph from the input set and apply graph convolutions~\cite{Simonovsky2017, Wang2019, Zhang2019, Liu2019DPAM, Shen2018, Dominguez2018, Chen_2020_CVPR, Rgcnn, Feng2019HGNN, Wang2018a, Zhang2018, Pan2018}. DGCNN \cite{Wang2019} introduces the EdgeConv operator and a dynamic graph component, which reconnects the k-nearest neighbor graph inside the network. EdgeConv computes the maximum over feature differences, which allows the network to represent directions in its channels. Channel-wise directions \textit{can} resemble spatial directions if spatial coordinates are provided as input, which is only the case in the first layer for DGCNN. In contrast, our convolutions support anisotropy directly in the operators.

\paragraph*{Rotation-equivariant approaches}
Architectures with two streams and vector-valued features are also used in rotation-equivariant approaches for point clouds and meshes. A group of works studies rotation-equivariance in 3D space, aiming to design networks invariant to rigid point-cloud transformations~\cite{Esteves2017, Thomas2018, Cohen2018, Poulenard2019SPHNet}. This concept is also incorporated in the transformer setups~\cite{fuchs2020se3transformer}. Rotation-equivariant kernels typically output vector-valued features. Vector Neurons simplify their use by linearly combining 3D vectors, followed by a vector non-linearity~\cite{deng2021vectorneurons}. Our use of vector MLPs is similar. Differences are that we use tangential vectors, rather than 3D vectors, and we derive these vectors inside the network using geometric operators.

An alternative approach is to build networks using intrinsic rotation-equivariant convolutions on meshes~\cite{dehaan2020gauge, Wiersma2020, CohenICML2019, Poulenard2018, weiler2021coordinate, gerken2021geometric}. These networks use local parametrizations and apply rotation- or gauge-equivariant kernels in the parameter domain to achieve independence from the choice of bases in the tangent spaces. Our approach is an alternative to gauge-equivariant networks. The use of differential operators also makes our networks independent of the choice of local coordinate systems. A benefit of our approach is that local parametrizations are not needed. 
For example, gauge-equivariant approaches typically use the exponential map for local parametrization but neglect the angular distortion induced by the parametrization.
To the best of our knowledge, we are the first to implement and evaluate an intrinsic two-stream architecture on point clouds.
 
\paragraph*{Geometric operators} 
Multiple authors use geometric operators to construct convolutions. The graph-Laplacian is used in GCN~\cite{kipf2017}. Spectral networks for learning on graphs are based on the eigenpairs of the graph-Laplacian~\cite{Bruna2013}. 
Surface networks for triangle meshes~\cite{Kostrikov2018} interleave the Laplacian with the extrinsic Dirac operator~\cite{liu2017dirac}.
Parametrized Differential Operators (PDOs)~\cite{jiang2018spherical} use the gradient and divergence operators to learn from spherical signals on unstructured grids.
DiffGCN~\cite{Eliasof2020} uses finite difference schemes of the gradient and divergence operators for the construction of graph networks.
DiffusionNet~\cite{Sharp2020DiffusionIA} learns diffusion using the Laplace--Beltrami operator and directional features from gradients. DeltaConv uses a larger set of operators, combining and concatenating operators from vector calculus. In addition, it allows the processing of directional information in the stream of vector-valued features. A related approach is HodgeNet~\cite{smirnov2021hodgenet}, which learns to build operators using the structure of differential operators.
Outside of deep learning, differential operators are widely applied for the analysis of 3D shapes \cite{Crane2013DGP,Goes2015}.

\section{Method}
We construct anisotropic convolutions by learning combinations of geometric differential operators. Because these operators are defined on scalar- and vector fields, we split our network into scalar and vector features. In this section, we describe these two streams, the operators and how they are discretized, and how combinations of the operators are learned. Finally, we consider the properties that result from this construction.

\paragraph*{Streams} Consider a point cloud $\mathbf{P} \in \mathbb{R}^{N \times 3}$ with $N$ points arranged in an $N \times 3$ matrix. All points can be associated with $C$ additional features, which are stored in a matrix $\mathbf{X} \in \mathbb{R}^{N \times C}$. Inside the network, we refer to the features in layer $l$ at point $i$ as $\mathbf{x}_i^{(l)} \in \mathbb{R}^{C_l}$. All of these features constitute the \textit{scalar stream}.

The \textit{vector stream} runs alongside the scalar stream. Each feature in the vector stream is a tangent vector, encoded by coefficients $(\alpha_i^u, \alpha_i^v)$ with respect to a basis in the corresponding tangent plane.
The basis can be any pair of orthonormal vectors that are orthogonal to the normal vector.
The coefficients are interleaved for each point, forming the matrix of features $\mathbf{V}^{(l)} \in \mathbb{R}^{2N \times C_l}$. One channel in $\mathbf{V}^{(l)}$ is a column of coefficients: $[ \alpha_1^u, \alpha_1^v, \ldots, \alpha_i^u, \alpha_i^v, \ldots, \alpha_N^u, \alpha_N^v ]^\intercal$. The input for the vector stream is a vector field defined at each point. In our experiments, we use the gradients of the input to the scalar stream. We will refer to the continuous counterparts of $\mathbf{X}$ and $\mathbf{V}$ as $X$ and $V$, respectively.

\subsection{Scalar to scalar: maximum aggregation}
A simplified version of point-based MLPs is applied inside the scalar stream, building on PointNet++ \cite{Qi2017b} and EdgeConv \cite{Wang2019}. We apply an MLP per point and then perform maximum aggregation over a $k$-nn neighborhood $\mathcal{N}(i)$. The features in the scalar stream are computed as
\begin{equation}
\mathbf{x}^{(l+1)}_i = h_{\mathbf{\Theta_0}}(\mathbf{x}^{(l)}_i) + \max_{j \in \mathcal{N}(i)}
h_{\mathbf{\Theta_1}}(\mathbf{x}^{(l)}_j),
\label{eq:scalarbase}
\end{equation}
where $h_{\mathbf{\Theta_0}}$ and $h_{\mathbf{\Theta_1}}$ denote multi-layer perceptrons (MLPs), consisting of fully connected layers, batch normalization~\cite{Ioffe2015}, and non-linearities. If point positions are used as input, they are centralized before maximum aggregation: $\mathbf{\hat{p}}_j = \mathbf{p}_j - \mathbf{p}_i$.

The biggest difference with EdgeConv and PointNet++ is that we use only point-based features within the network instead of edge-based features. The matrix multiplication used inside the MLP is thus not applied to $kN$ feature vectors, but $N$ point-wise feature vectors. This has a significant impact on the run time of the forward and backward passes. Directional information is encoded in per-point vectors instead of edges.

\subsection{Scalar to vector: Gradient and co-gradient}
The gradient and co-gradient operators connect the scalar stream to the vector stream. The gradients of a function represent the largest rate of change and the directions of that change as a vector at each point.
The co-gradients are 90-degree rotations of the gradients. Combined, the gradients and co-gradients span the tangent planes, allowing the network to scale, skew, and rotate the gradient vectors.

We construct a discrete gradient operator using a moving least-squares approach on neighborhoods with $k$ neighbors~\cite{Nealen2004}. This approach is used in modeling and processing for point clouds and solving differential equations on point clouds~\cite{crane2013geodesics, Liang2013SolvingPD}. The procedure and accompanying theory is outlined in the supplemental material. The gradient operator is represented as a sparse matrix $\mathbf{G} \in \mathbb{R}^{2N \times N}$. It takes $N$ values representing features on the points and outputs $2N$ values representing the gradient expressed in coefficients of the tangent basis of each point. The matrix is highly sparse as it only contains $2k$ elements in each row. The co-gradient $\mathbf{JG}$ is a composition of the gradient with a block-diagonal sparse matrix $\mathbf{J} \in \mathbb{R}^{2N \times 2N}$, where each block in $\mathbf{J}$ is a $2\times2 $ 90-degree rotation matrix.

\begin{figure}[b]
    \centering
        \includegraphics[width=0.4\columnwidth]{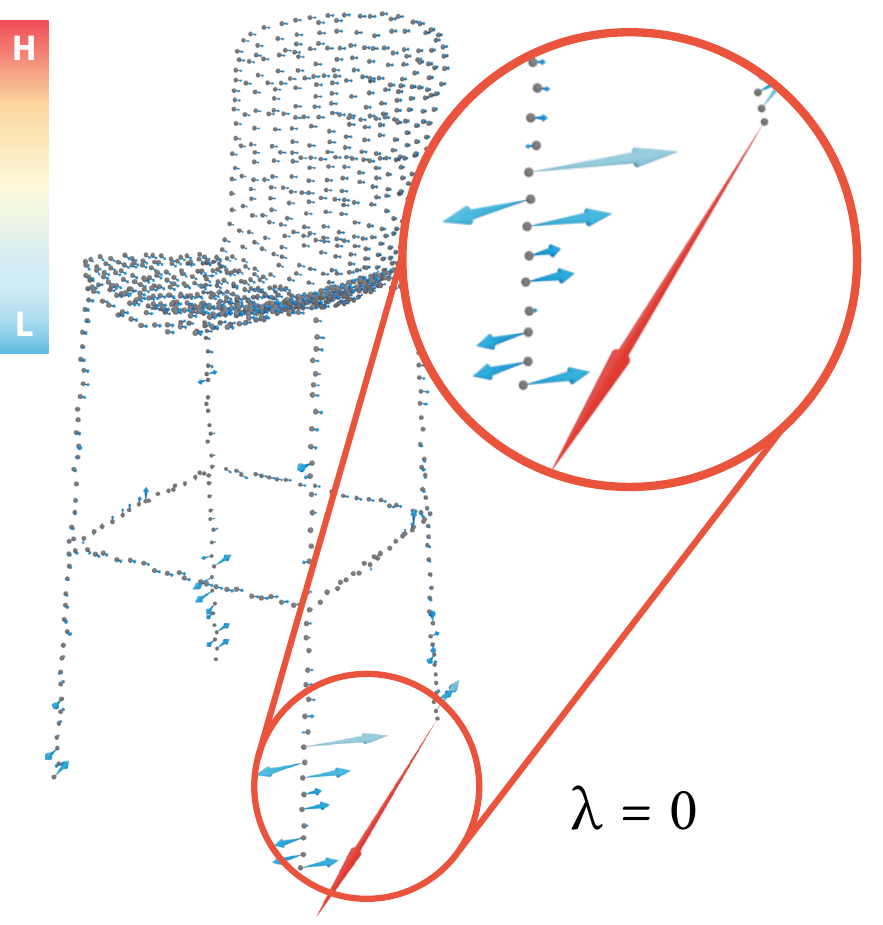}
        \includegraphics[width=0.4\columnwidth]{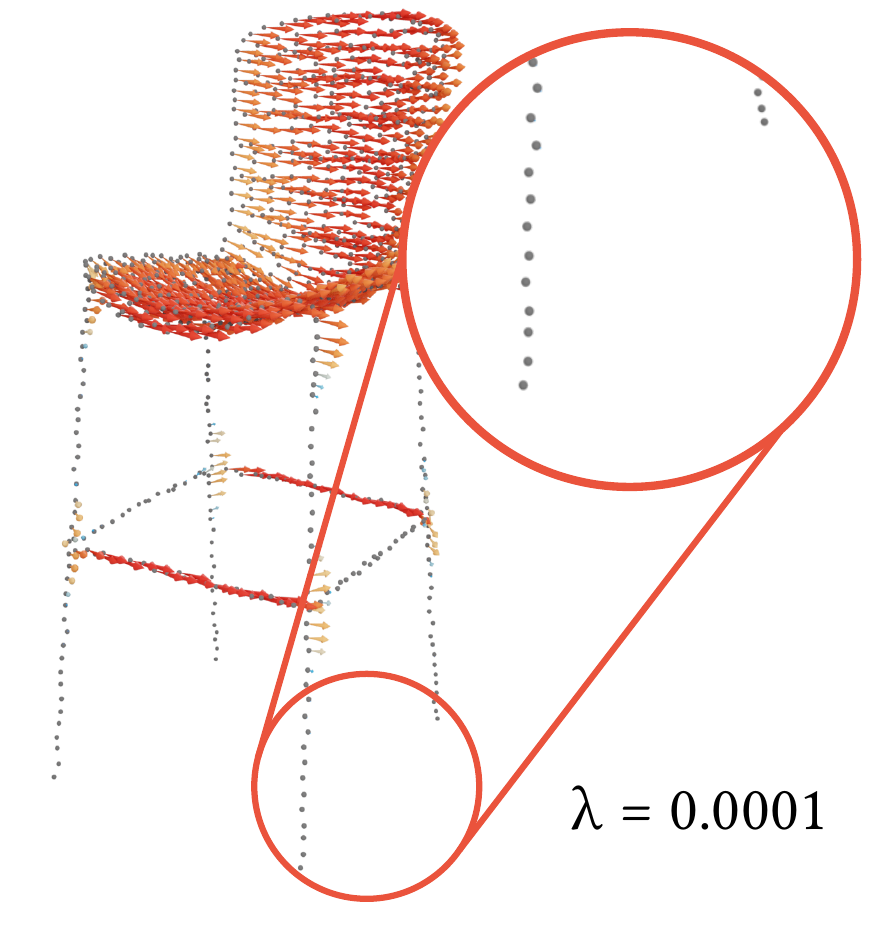}
    \caption{Gradient of the x-coordinate on a chair without regularization (left) and with regularization (right).
    }
    \label{fig:chairregularizer}
\end{figure}

Point clouds typically contain undersampled regions and noise. This can be problematic for the moving least-squares procedure. Consider the example in \autoref{fig:chairregularizer}, a chair with thin legs. Only a few points lie along the line constituting the legs of the chair. Hence, the perpendicular direction to the line is undersampled, resulting in a volatile least-squares fit: a minor perturbation of one of the points can heavily influence the outcome (left, circled area). We add a regularization term scaled by $\lambda$ to the least-squares fitting procedure, which seeks to mitigate this effect (right). This is a known technique referred to as ridge regression or Tikhonov regularization.

We also argue that the gradient operator should be normalized, motivated by how information is fused in the network. If $\mathbf{G}$ exhibits diverging or converging behavior, features resulting from $\mathbf{G}$ will also diverge or converge. This is undesirable when the gradient is applied multiple times in the network. Features arising from the gradient operation would then have a different order of magnitude which needs to be accounted for by the network weights. Therefore, we normalize $\mathbf{G}$ by the  $\ell_\infty$-operator norm, which provides an upper bound on the scaling behavior of an operator
\begin{equation}
    \mathbf{\hat{G}} = \mathbf{G} / |\mathbf{G}|_{\infty},\, \quad\text{where} \,\, |\mathbf{G}|_{\infty} = \max_{i} \sum_{j} |\mathbf{G}_{ij}|.
\end{equation}

\subsection{Vector to scalar: Divergence, Curl, and Norm}
The vector stream connects back to the scalar stream with divergence, curl, and norm. These operators are commonly used to analyze vector fields and indicate features such as sinks, sources, vortices, and the strength of the vector field. The network can use them as building blocks for anisotropic operators.

The discrete divergence is also constructed with a moving least-squares approach, which is described in the supplement. Divergence is represented as a sparse matrix $\mathbf{D} \in \mathbb{R}^{N \times 2N}$, with $2kN$ elements. Curl is derived as $-\mathbf{DJ}$.

\subsection{Vector to vector: Hodge Laplacian}
Vector features are diffused in the vector stream using a combination of the identity $\mathbf{I}$ and the Hodge Laplacian $\bm{\Delta}$ of $V$.
Applying the Hodge Laplacian to a vector field $V$ results in another vector field encoding the difference between the vector at each point and its neighbors. The Hodge Laplacian can be formulated as a combination of grad, div, curl and $\mathcal{J}$~\cite{Brandt2017}
\begin{equation}
    \bm{\Delta} = -(\text{grad\,} \text{div\,} + \mathcal{J}\text{\,grad\,} \text{curl\,}).
\end{equation}
In the discrete setting, we replace each operator with its discrete variant
\begin{equation}
    \mathbf{L} = -(\mathbf{GD} - \mathbf{JGDJ}).
\end{equation}

\subsection{Why these operators?}
The operators we use are related to each other in a fundamental way. They form a metric version of the \textit{de Rham complex} of a surface \cite{Wardetzky2006}. The following diagram lays out the connections described in the previous sections, where each of the operators maps between functions (scalar fields) and vector fields.
\begin{equation}
    \begin{tikzcd}
    X \arrow[rr, "\text{grad}", shift left]
        &
        & V \arrow[ll, "\text{div}", shift left]
            \arrow[rr, "\text{curl}", shift left]
            &
            & X \arrow[ll, "\text{co-grad}", shift left]
    \end{tikzcd}
\end{equation}
Note that the bottom row is a 90-degree rotated version of the top row. If we follow the diagram from left to right and apply grad and then curl to any function, the output will always be zero. The same holds for the path from right to left. The operators listed are first-order derivatives. Laplacians, which are second-order derivatives, can be formed by composing the first-order operators. For functions: to vector fields with grad and back again with div (Laplace-Beltrami). For vector fields: we go to scalars with div and curl and back again with grad and co-grad (Hodge-Laplacian). DeltaConv learns to combine these operators and supports anisotropy by adding non-linearities in-between.

\begin{figure}[t]
    \centering
    \includegraphics[width=0.9\columnwidth]{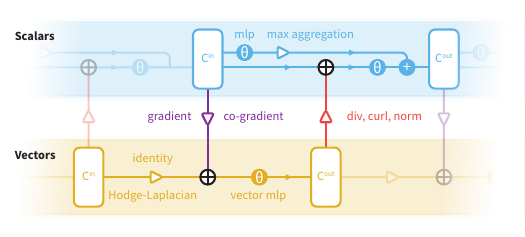}
    \caption{Schematic of DeltaConv.}
    \label{fig:streams}
\end{figure}

\subsection{DeltaConv: Learning Anisotropic Operators}
Each of the operations either outputs scalar-valued or vector-valued features. We concatenate all the features belonging to each stream and then combine these features with parametrized functions
\begin{align}
    \mathbf{v}_i' &= \mathbf{h}_{\mathbf{\Theta_0}}(\mathbf{v}_i, \,  (\mathbf{GX})_i\, ,  (\mathbf{LV})_i), \nonumber \\
    \mathbf{x}_i' &= h_{\mathbf{\Theta_1}}(\mathbf{x}_i, (\mathbf{DV'})_i\, , (-\mathbf{DJV'})_i\, , \Vert \mathbf{v'}_i\Vert) + \max_{j \in \mathcal{N}_i}\,h_{\mathbf{\Theta_2}}(\mathbf{x}_j). 
    \label{eq:fusingstreams}
\end{align}
We use the prime to indicate features in layer $l+1$. All other features are from layer $l$. $h_{\mathbf{\Theta_1}}$ and $h_{\mathbf{\Theta_2}}$ denote standard MLPs. $\mathbf{h}_{\mathbf{\Theta_0}}$ denotes an MLP used for vectors.
The vector MLPs scale and sum vectors, which means they do not work on individual vector coefficients and are coordinate-independent. Recall that $\mathbf{V} \in \mathbb{R}^{2N \times C^{(l)}}$ interleaves the vector coefficients for each point in the columns. One layer in the vector MLP is applied to $\mathbf{V}$ as follows
\begin{equation}
    \mathbf{V}' = \sigma(\mathbf{VW}),
\end{equation}
where $\mathbf{W} \in \mathbb{R}^{C^{(l)} \times C^{(l+1)}}$ is a weight matrix and $\sigma$ is a non-linearity applied to vector norms.
Matrix multiplication with $\mathbf{W}$ linearly combines the vector features but the individual coefficients of a vector are not mixed. Before the vector MLP is applied, we concatenate the 90-degree rotated vectors to the input features. This allows the MLP to also rotate vector features and enriches the set of operators. For example, the 90-degree rotated gradient is the co-gradient. The vector MLP can learn to combine information from local neighborhoods (through the gradient and Hodge--Laplacian), as well as information from different channels (through the identity). A schematic overview of \autoref{eq:fusingstreams} can be found in \autoref{fig:streams}.

While \autoref{eq:fusingstreams} formulates DeltaConv in terms of MLPs and feature concatenation, an alternative perspective is to consider the operations in \autoref{eq:fusingstreams} as linearly combining the elementary operators and composing them with non-linearities in-between to form anisotropic geometric operators.

\subsection{Properties of DeltaConv}
The building blocks of DeltaConv, such as the gradient, divergence, curl, and the combination with non-linearities allow DeltaConv to build nonlinear anisotropic convolution filters. This is illustrated by the example of the Perona--Malik filter in \autoref{fig:peronamalik}.
The vector stream also allows DeltaConv to process vector features and their relative directions directly with the appropriate operators.

DeltaConv is formulated in terms of smooth differential operators and is not restricted to a specific surface representation. In this work, we implement DeltaConv for point clouds and images. However, the concepts generalize to other representations. For example, an implementation for meshes could be done using finite element discretizations \cite{Brandt2017} or discrete exterior calculus \cite{Crane2013DGP}.

DeltaConv is coordinate-independent, meaning that the weights used in DeltaConv do not depend on the choices of tangent bases. For example, a forward pass on a shape with one choice of bases leads to the same output and weight updates when run with different bases.
The coordinate-independence follows from the fact that all elementary operations in DeltaConv, such as applying geometric operators and vector MLPs, are coordinate-independent. It is known from differential geometry that one obtains the same results with geometric operators, no matter which basis is chosen \cite{o1983semi}. This property is preserved by the discretization of the operators and thus inherited by DeltaConv.

Finally, each of the building blocks of DeltaConv is isometry invariant. That means DeltaConv does not change if a shape is isometrically deformed. This property can be beneficial for tasks where shapes are rigidly or non-rigidly deformed. If the surface orientation is flipped, rotations in the tangent plane are flipped as well. DeltaConv is robust to this if only the gradient and divergence are used.

\section{Experiments}
We validate our approach with comparisons to state-of-the-art approaches on classification and segmentation. In addition, we perform ablation studies to provide more insight into the effect of the vector stream on anisotropy, accuracy, and efficiency.

\subsection{Implementation details}
In our experiments we use network architectures based on DGCNN \cite{Wang2019}. We replace each EdgeConv block with a DeltaConv block (\autoref{fig:streams}) and do not use the dynamic graph component. Thus, the networks operate at a single scale on local neighborhoods. Despite this simple architecture, DeltaConv achieves state-of-the-art results. To show what architectural optimizations mean for DeltaConv, we also test the U-ResNet architecture used in KPFCNN~\cite{thomas2019KPConv} but with the convolution blocks in the encoder replaced by DeltaConv blocks. In the downsampling blocks used by these networks, we pool vector features by averaging them with parallel transport \cite{Wiersma2020}. More details are provided in the supplemental material. Code is available at \url{https://github.com/rubenwiersma/deltaconv}.

\emph{Data transforms.} A $k$-nn graph is computed for every shape. This graph is used for maximum aggregation in the scalar stream. It is reused to estimate normals when necessary and to construct the gradient. For each experiment, we use xyz-coordinates as input to the network and augment them with a random scale and translation, similar to previous works. Some datasets require specific augmentations, which are detailed in their respective sections.

\emph{Training.} The parameters in the networks are optimized with stochastic gradient descent (SGD) with an initial learning rate of $0.1$, momentum of $0.9$ and weight decay of $0.0001$. The learning rate is updated using a cosine annealing scheduler~\cite{cosineannealing2017}, which decreases the learning rate to $0.001$.

\subsection{Classification}
For classification, we study ModelNet40 \cite{Wu2015}, ScanObjectNN \cite{uy-scanobjectnn-iccv19}, and SHREC11 \cite{lian2011}. With these experiments, we aim to demonstrate that our networks can achieve state-of-the-art performance on a wide range of challenges: point clouds sampled from CAD models, real-world scans, and non-rigid, deformable objects.

\begin{table}[t]
    \caption{Classification results on ModelNet40.}
    \label{table:modelnet}
    \begin{center}
        \resizebox{0.9\columnwidth}{!}{
            \begin{tabular}{lcr}
                \toprule
                Method                                                    & Mean           & Overall       \\
                                                                          & Class Accuracy & Accuracy      \\
                \midrule
                PointNet++ \cite{Qi2017b}                                 & -              & 90.7          \\
                PointCNN \cite{li2018pointcnn}
                                                                          & 88.1           & 92.2          \\
                DGCNN \cite{Wang2019}                                     & 90.2           & 92.9          \\
                KPConv deform \cite{thomas2019KPConv}                     & -              & 92.7          \\
                KPConv rigid \cite{thomas2019KPConv}                      & -              & 92.9          \\
                DensePoint \cite{Densepoint2019}                          & -              & 93.2          \\
                RS-CNN \cite{liu2019rscnn}                                & -              & 93.6          \\
                GBNet \cite{gbnet2021shi}                                        & 91.0              & \textbf{93.8}       \\
                PointTransformer \cite{pointtransformer}                  & 90.6             & 93.7          \\
                PAConv \cite{paconv}                          & -              & 93.6          \\
                Simpleview \cite{simpleview}                              & -              & 93.6          \\
                Point Voxel Transformer \cite{pointvoxeltransformer}      & -              & 93.6          \\
                CurveNet \cite{curvenet}                      & -              & \textbf{93.8}          \\
                \midrule
                DeltaNet (ours)                                           & \textbf{91.2}  & \textbf{93.8} \\
                \bottomrule
            \end{tabular}
        }
    \end{center}
\end{table}

\paragraph{ModelNet40}
The ModelNet40 dataset \cite{Wu2015} consists of 12,311 CAD models from 40 categories. 9,843 models are used for training and 2,468 models for testing. Each point cloud consists of 1,024 points sampled from the surface using a uniform sampling of 8,192 points from mesh faces and subsequent furthest point sampling (FPS). We use 20 neighbors for maximum aggregation and to construct the gradient and divergence. Ground-truth normals are used to define tangent spaces for these operators and the regularizer is set to $\lambda=0.01$. As input to the network, we use the xyz-coordinates. The classification architecture is optimized for 250 epochs. We do not use any voting procedure and list results without voting.

The results for this experiment can be found in \autoref{table:modelnet}. DeltaConv improves significantly on the most related maximum aggregation operators and is on par with or better than state-of-the-art approaches.

\paragraph{ScanObjectNN}
ScanObjectNN~\cite{uy-scanobjectnn-iccv19} contains 2,902 unique object instances with 15 object categories sampled from SceneNN~\cite{scenenn-3dv16} and ScanNet~\cite{dai2017scannet}. The dataset is enriched to $\sim15,000$ objects by preserving or removing background points and by perturbing bounding boxes. The variant without background points is tested without any perturbations (\textsc{no bg}). The variant with background points is both tested without (\textsc{bg}) and with perturbations: Bounding boxes are translated (\textsc{t}), rotated (\textsc{r}), and scaled (\textsc{s}) before each shape is extracted. This means that some shapes are cut off, rotated, or scaled. \textsc{t25} and \textsc{t50} denote a translation by 25\% and 50\% of the bounding box size, respectively.

We use a modified version of the classification architecture with four convolution blocks with the following output dimensions: 64, 64, 64, 128. This setup matches the architecture used for DGCNN in \cite{uy-scanobjectnn-iccv19}. Normals are estimated with 10 neighbors per point and the operators are constructed with 20 neighbors and $\lambda = 0.001$. As input, we provide the xyz-positions, augmented with a random rotation around the up-axis and a random scale $S \in \mathcal{U}(4/5, 5/4)$. The network is trained for 250 epochs.

Our results are compared to those reported by the authors of ScanObjectNN (row 1-8)~\cite{uy-scanobjectnn-iccv19} and other recent approaches in \autoref{table:scanobjectnn}. We find that our approach outperforms all networks for every type of perturbation, including networks that explicitly account for background points.

\begin{table}[t]
   \caption{Classification results on ScanObjectNN.}
   \label{table:scanobjectnn}
   \begin{center}
      \resizebox{\columnwidth}{!}{
         \begin{tabular}{l|c|ccccr}
            \toprule
            Method                                        & \textsc{no bg} & \textsc{bg}   & \textsc{t25}  & \textsc{t25r} & \textsc{t50r} & \textsc{t50rs} \\
            \midrule
            3DmFV \cite{ben20183dmfv}               & 73.8           & 68.2          & 67.1          & 67.4          & 63.5          & 63.0           \\
            PointNet \cite{Qi2017a}                 & 79.2           & 73.3          & 73.5          & 72.7          & 68.2          & 68.2           \\
            SpiderCNN \cite{xu2018spidercnn}        & 79.5           & 77.1          & 78.1          & 77.7          & 73.8          & 73.7           \\
            PointNet++ \cite{Qi2017b}               & 84.3           & 82.3          & 82.7          & 81.4          & 79.1          & 77.9           \\
            DGCNN \cite{Wang2019}                   & 86.2           & 82.8          & 83.3          & 81.5          & 80.0          & 78.1           \\
            PointCNN \cite{li2018pointcnn}          & 85.5           & 86.1          & 83.6          & 82.5          & 78.5          & 78.5           \\
            BGA-PN++ \cite{uy-scanobjectnn-iccv19}  & -              & -             & -             & -             & -             & 80.2           \\
            BGA-DGCNN \cite{uy-scanobjectnn-iccv19} & -              & -             & -             & -             & -             & 79.9           \\
            GBNet \cite{gbnet2021shi}                      & -              & -             & -             & -             & -             & 80.5           \\
            GDANet \cite{gdanet}                    & 88.5           & 87.0          & -             & -             & -             & -           \\
            DRNet \cite{drnet}                      & -              & -             & -             & -             & -             & 80.3           \\
            \midrule
            DeltaNet (ours)                         & \textbf{89.5}  & \textbf{89.3} & \textbf{89.4} & \textbf{87.0} & \textbf{85.1} & \textbf{84.7}  \\
            \bottomrule
         \end{tabular}
      }
   \end{center}
\end{table}
\begin{table}[b]
    \caption{Classification results on SHREC11.}
    \label{tab:shrec11-class}
    \begin{center}
        \resizebox{0.8\columnwidth}{!}{
            \begin{tabular}{lr}
                \toprule
                Method                                & Accuracy      \\ \midrule
                MeshCNN \cite{hanocka2019meshcnn}     & 91.0          \\
                HSN \cite{Wiersma2020}                & 96.1          \\
                MeshWalker \cite{Lahav2020MeshWalker} & 97.1          \\
                PD-MeshNet \cite{milano2020pd}        & 99.1          \\
                HodgeNet \cite{smirnov2021hodgenet}   & 94.7          \\
                FC \cite{mitchel2021field}            & 99.2          \\
                DiffusionNet (xyz) \cite{Sharp2020DiffusionIA} & 99.4          \\
                DiffusionNet (hks) \cite{Sharp2020DiffusionIA} & 99.5  \\
                \midrule
                DeltaNet (ours)                       & \textbf{99.6} \\
                \bottomrule
            \end{tabular}
        }
    \end{center}
\end{table}

\paragraph{SHREC11}
The SHREC11 dataset \cite{lian2011} consists of 900 non-rigidly deformed shapes, 30 each from 30 shape classes. This experiment aims to validate the claim that our approach is well suited for non-rigid deformations. Like previous works \cite{hanocka2019meshcnn, Wiersma2020, Sharp2020DiffusionIA}, we train on 10 randomly selected shapes from each class and report the average over 10 runs. We sample 2048 points from the simplified meshes used in MeshCNNs experiments~\cite{hanocka2019meshcnn} and use 20 neighbors and mesh normals to construct the operators ($\lambda = 0.001$). As input, we provide xyz-coordinates, which are randomly rotated along each axis. We decrease the number of parameters in each convolution of the classification architecture to 32, since the dataset is much smaller than other datasets. The network is trained for 100 epochs. We find that our architecture is able to improve on state-of-the-art results (\autoref{tab:shrec11-class}), validating the effectiveness of our intrinsic approach on deformable shapes.

\subsection{Segmentation}
For segmentation, we evaluate our architecture on ShapeNet (part segmentation) \cite{yi2016}.
ShapeNet consists of 16,881 shapes from 16 categories. Each shape is annotated with up to six parts, totaling 50 parts. We use the point sampling of 2,048 points provided by the authors of PointNet~\cite{Qi2017a} and the train/validation/test split follows \cite{Chang2015ShapeNet}. The operators are constructed with 30 neighbors and ground-truth normals to define tangent spaces ($\lambda=0.001$). The xyz-coordinates are provided as input to the network, which is trained for 200 epochs. During testing, we evaluate each shape with ten random augmentations and aggregate the results with a voting procedure. Such a voting approach is used in the most recent works that we compare with.

\begin{figure}[t]
    \centering
    \includegraphics[width=0.85\columnwidth]{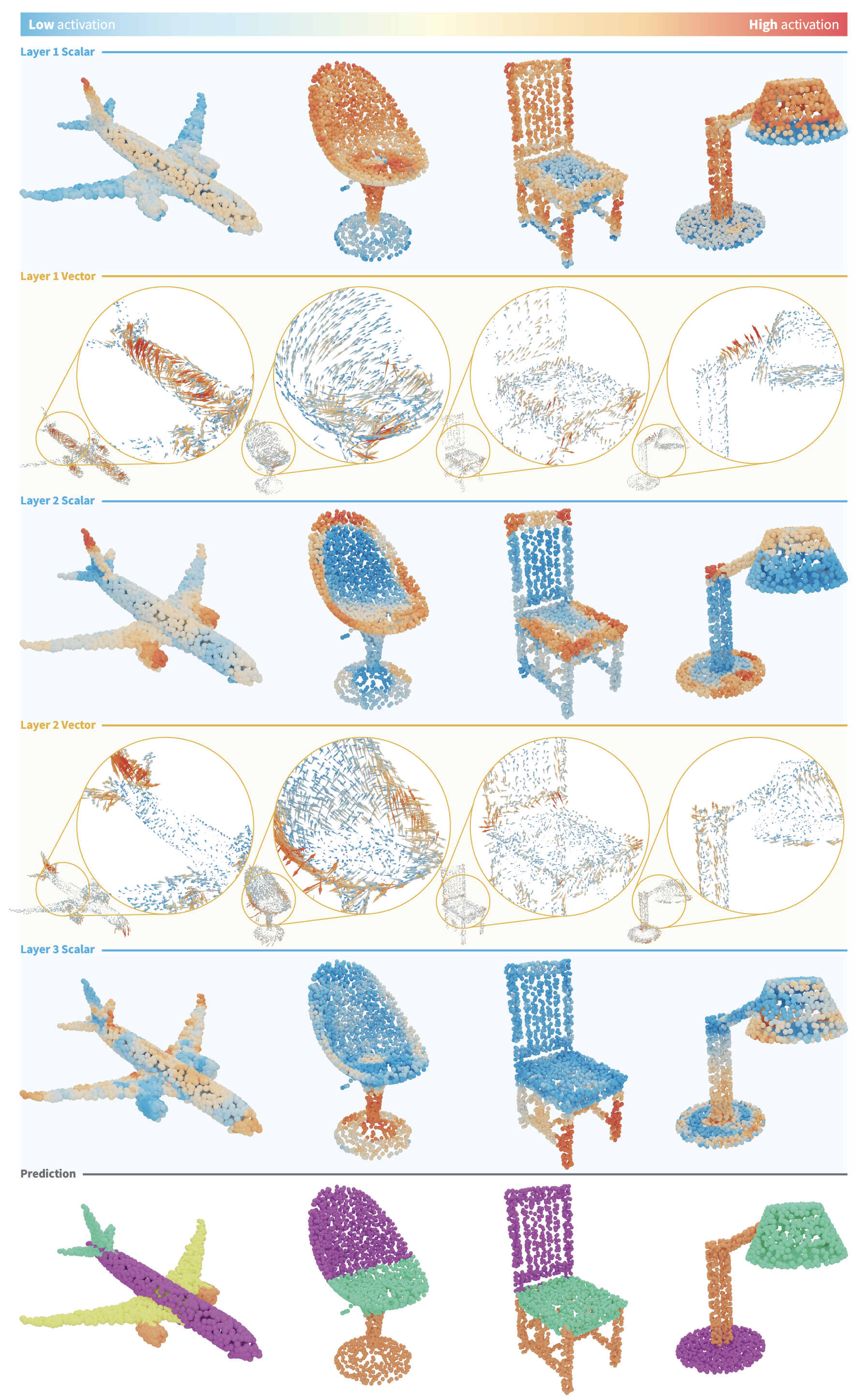}
    \caption{For each layer of the network, we show how a single scalar- or vector-feature varies over shapes in ShapeNet. The last row shows the output of the network. The features tend to activate on similar regions.}
    \label{fig:featurevis}
\end{figure}

The results are shown in \autoref{table:part_segmentation}, where our approach, especially the U-ResNet variant, improves upon the state-of-the-art approaches on the mean instance IoU metric and in many of the shape categories (full breakdown in the supplemental material). For each category, DeltaConv is either comparable to or better than other architectures and significantly better than the most related intrinsic approaches (PointNet++ and DGCNN). In \autoref{fig:featurevis}, we provide feature visualizations to give an idea of the features derived by the network.

\begin{table}[t]
    \caption{Part segmentation results on ShapeNet.}
    \label{table:part_segmentation}
    \begin{center}
        \resizebox{0.8\columnwidth}{!}{
            \begin{tabular}{lr}
                \toprule
                Method          & Mean   \\
                                & inst. mIoU \\
                \midrule
                PointNet++ \cite{Qi2017b}      & 85.1                \\
                PointCNN \cite{li2018pointcnn}        & 86.1                \\
                DGCNN \cite{Wang2019}           & 85.2                \\
                KPConv deform \cite{thomas2019KPConv}   & 86.4                \\
                KPConv rigid \cite{thomas2019KPConv}    & 86.2                \\
                GDANet \cite{gdanet}          & 86.5                \\
                PointTransformer \cite{pointtransformer} & 86.6               \\
                PointVoxelTransformer \cite{pointvoxeltransformer} & 86.5          \\
                CurveNet \cite{curvenet}        & 86.8  \\
                \midrule
                DeltaNet (ours) &  86.6               \\
                Delta-U-ResNet (ours) & \textbf{86.9} \\
                \bottomrule
                \end{tabular}
        }
    \end{center}
\end{table}

\subsection{Ablation Studies}
We aim to validate the claim of anisotropy, isolate the effect of the vector stream, validate the choices to regularize and normalize the gradient and divergence operators, and investigate the impact of our approach on the timing and parameter counts of these networks.

\paragraph*{Anisotropy} To validate that DeltaConv supports anisotropy, we train a network to mimic anisotropic diffusion~\cite{peronamalik}. A ResNet~\cite{he2016deep} with 16 layers and 16 channels in the hidden layers is trained for 100 iterations with Adam \cite{adamkingma} to match a target image generated with 20 anisotropic diffusion steps. In each diffusion step, the gradients are scaled with $\exp(-(|v|/0.05)^2)$. We vary the convolution blocks in the network with the ones from DiffusionNet~\cite{Sharp2020DiffusionIA}, EdgeConv~\cite{Wang2019}, PointNet++~\cite{Qi2017b}, GCN~\cite{kipf2017}, and regular image CNNs. For DiffusionNet, we set the diffusion time to a fixed value, as we are interested in the ability of the convolution to derive anisotropic filters through its gradient features. For all other convolutions, the neighborhoods are 3x3 pixel blocks. The results are shown in \autoref{fig:peronamalik} and in the supplement. DeltaConv achieves a good match. The other operators tend to blur the image or produce artifacts. For PointNet and EdgeConv, this is likely due to the variable nature and sharpness of the maximum aggregation. DiffusionNet lacks the divergence and curl operators and does not maintain a vector stream, which is necessary to analyze the relative directions of vector features in local neighborhoods.

\paragraph*{Effectiveness of vector stream} To study the benefit of the vector stream and its effect on different types of intrinsic scalar convolutions, we set up three different scalar streams: (1) a Laplace--Beltrami operator, $\Delta = -\text{div}\,\text{grad}$, (2) GCN \cite{kipf2017}, and (3) maximum aggregation (\autoref{eq:scalarbase}). We test three variants of each network: (1) only scalar stream, (2) scalar stream with the number of parameters adjusted to match a two-stream architecture, and (3) both the scalar and vector stream.

We test each configuration on ModelNet40 and ShapeNet. For both of these tasks, we use the DGCNN base architecture. The model for ShapeNet is trained for 50 epochs to save on training time and no voting is used, which results in slightly lower results than listed in \autoref{table:part_segmentation}. The results are listed in \autoref{table:cls_abl}. We find that the vector stream improves the network for each scalar stream for both tasks, reducing the error between $19-25\%$ for classification and $3-21\%$ for segmentation. For maximum aggregation on ShapeNet, the improvements are lower, but still considerable, given the rate of progress on this dataset over the last few years. Simply increasing the number of parameters in the scalar stream does not yield the same improvement as adding the vector stream, showing that the vector-valued features are of meaningful benefit. Maximum aggregation in the scalar stream yields the highest accuracy.

\begin{table}[t]
    \caption{Ablations of DeltaConv on ShapeNet (Seg) and ModelNet40 (M40) with varying scalar streams.}
    \label{table:cls_abl}
    \begin{center}
        \resizebox{\columnwidth}{!}{
            \begin{tabular}{lccccc}
                \toprule
                Scalar          & Vector                    & Match                     & Seg mIoU                     & M40 mcA                        & M40 OA                         \\
                Convolution        & Stream                      & $\#$ params & & & \\ \midrule            
Laplace--       & -                         & -                         & 82.5                           & 86.1                           & 90.4                           \\ 
Beltrami        & -                         & \checkmark & 82.5                           & 87.1                           & 90.6                           \\ 
                & \checkmark & -                         & \textbf{84.9} & \textbf{89.4} & \textbf{92.2} \\ \midrule
GCN             & -                         & -                         & 81.1                           & 87.3                           & 90.4                           \\
                & -                         & \checkmark & 81.2                           & 87.3                           & 90.8                           \\ 
                & \checkmark & -                         & \textbf{85.1} & \textbf{90.6} & \textbf{92.8} \\ \midrule
Max aggregation & -                         & -                         & 85.7                           & 89.2                           & 92.2                           \\  
                & -                         & \checkmark & 85.7                           & 89.5                           & 92.6                           \\
                & \checkmark & -                         & \textbf{86.1} & \textbf{91.2} & \textbf{93.8} \\
                \bottomrule
            \end{tabular}
        }
    \end{center}
\end{table}
\begin{table}[b]
    \caption{Timing and parameter counts for classification on ModelNet40. The timing for training and inference includes all necessary precomputations.}
    \label{table:timings}
    \begin{center}
        \resizebox{\columnwidth}{!}{
            \begin{tabular}{lccccr}
                \toprule
                Convolution                         & Data            & Training         & Backward & Inference & $\#$ Params \\
                                                    & Transform      &                 &           & &  \\
                \midrule
                DeltaConv (Lapl.)              & k-nn + ops   & $80$ms         & $5$ms &$80$ms          & 2,036,938        \\
                DeltaConv                           & k-nn + ops   & $130$ms   & $60$ms & $125$ms       & 2,037,962        \\
                EdgeConv                            & k-nn          & $196$ms   & $147$ms & $186$ms      & 1,801,610        \\
                \bottomrule
            \end{tabular}
        }
    \end{center}
\end{table}

\paragraph*{Timing and parameters} In our method section we argue that computing the gradient matrix is lightweight and that the simplified maximum aggregation operator is significantly faster than edge-based operators in PointNet++ and DGCNN. The main bottleneck in these convolutions is maximum aggregation over each edge. In this experiment, we demonstrate this by reporting the time it takes to train and test the classification network on one batch of 32 shapes with 1,024 points each. This includes all precomputation steps, such as computing the k-nearest neighbor graph ($\sim15$ms) and constructing the gradient and divergence operators ($\sim30$ms). The EdgeConv network is tested without a dynamic graph component, so that only the effect of precomputation and convolutions remains. All timings are obtained on the same machine with an NVIDIA RTX 2080Ti after a warm-up of 10 iterations. We implemented each method in PyTorch~\cite{pytorch} and PyTorch Geometric~\cite{Fey/Lenssen/2019}.
The results are listed in \autoref{table:timings}. We find that our network only increases the number of parameters by $10\%$. Our network is significantly faster than the edge-based convolution: $1.5\times$ faster in training and inference and $2.5\times$ faster in the backward pass. DeltaConv with a Laplacian in the scalar stream is even faster: $>2\times$ faster in training and inference and $30\times$ faster in the backward pass.

\paragraph*{Gradient regularization and normalization}
In our method section, we argue that the least-squares fit for constructing the gradient and divergence should be regularized and the operators should be normalized. In this experiment, we intend to validate these choices. We train a model that is entirely based on our gradient operator, with a Laplace--Beltrami operator in the scalar stream. This means that every spatial operator in the network is influenced by regularization and scaling. The model is trained on the ModelNet40 for 50 epochs. The results are listed in \autoref{table:regularizer_scaling_abl}. We notice a considerable difference between our approach with- and without regularization. There is a $2.8$ percentage point decrease in mean class accuracy and $1.7$ percentage point decrease in overall accuracy when the operator is not normalized.

\begin{table}[t]
    \caption{Classification accuracy on ModelNet40 with and without regularization and normalization.}
    \label{table:regularizer_scaling_abl}
    \begin{center}
        \resizebox{0.8\columnwidth}{!}{
            \begin{tabular}{lccr}
                \toprule
                $\lambda$ & Normalization & Mean           & Overall       \\
                        &        & Class Accuracy & Accuracy      \\
                \midrule
                $10^{-32}$ & $\checkmark$ & 85.2 & 90.3 \\
                $10^{-2}$ & - &  86.6 & 90.5 \\
                $10^{-2}$ & $\checkmark$ & \textbf{89.4} & \textbf{92.2} \\
                \bottomrule
            \end{tabular}
        }
    \end{center}
\end{table}

\section{Conclusion}
In this work, we propose DeltaConv, a new convolutional layer for point cloud CNNs that is capable of extracting and processing directional features. DeltaConv separates features into a scalar- and vector stream and uses linear combinations and compositions of a selected set of geometric operators from vector calculus to map between and along the streams. 
This construction allows DeltaConv networks to learn anisotropic convolutions fitting to the data and task at hand.
We demonstrate improved performance on a wide range of tasks, showing the potential of using DeltaConv in a learning setting on point clouds. We hope that this work will provide insight into the functionality and operation of neural networks for point clouds and spark more work that combines learning approaches with powerful tools from geometry processing.

\paragraph*{Challenges and future work}
We limit our study to analysis tasks. While we do not think it is impossible to adapt our operators for generative tasks, it is unclear if and when the operators should be recomputed when a surface is generated. Our work opens up interesting possibilities for future work. Besides exploring more applications of the vector stream, we want to test our approach on other surface discretizations and other manifolds (e.g., hyperbolic spaces and higher dimensional spaces) for which these operators are available, and also intend to study how other variants of the scalar stream impact the network.
\begin{acks}
We thank the anonymous reviewers for their constructive feedback. This work has been partially supported by the NWO VIDI grant NextView, a gift from the TU Delft Universiteitsfonds, and a gift from Google Cloud. Ahmad Nasikun was supported through a doctoral scholarship from the Indonesia Endowment Fund for Education (LPDP).
\end{acks}

\bibliographystyle{ACM-Reference-Format}
\bibliography{main}

\appendix
\begin{figure*}[t]
    \includegraphics[width=\textwidth]{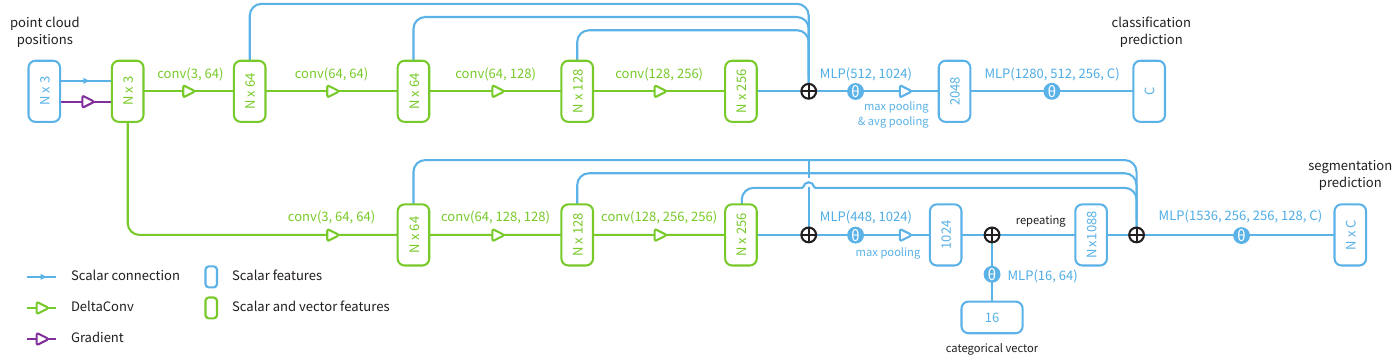}
    \caption{The two architectures used for classification and segmentation, based on \cite{Wang2019}. Please refer to Equation 6 and Figure 4 in the main text for the formulation of each convolution and how the streams are combined.}
    \label{fig:architecture}
    \centering
\end{figure*}

\section{Discretized Operators}
\paragraph{Gradient}
We construct a discrete gradient using the moving least-squares approach from~\cite{Liang2013SolvingPD}. We go through each step to show how to derive the gradient starting with the general formula from Riemannian geometry and simplify terms whenever the setting allows us to do so. 

We locally fit a surface patch to estimate the metric at each point $p$ using moving least-squares~\cite{Nealen2004}.
The surface patch $\g : \mathbf{\Omega} \subset \mathbb{R}^2 \rightarrow \mathbb{R}^3$, often called a Monge patch, describes the surface as a quadratic polynomial $h(u,v)$ over the tangent plane at $p$ and is given by \begin{equation}
    \g(u, v) = \left[u, v, h(u, v)\right]^\intercal,
\end{equation} 
where $u, v$ denote local coordinates in the tangent plane. Since the surface patch should interpolate the point $p$ and the surface normal of the patch at $p$ should agree with the normal of the tangent plane at $p$, the constant and linear terms of $h(u,v)$ vanish
\begin{align}
    h(u, v) &= \alpha_1 u^2 + \alpha_2 uv + \alpha_3 v^2, \\
    h_u &= 2\alpha_1 u + \alpha_2 v, \\
    h_v &= \alpha_2 u + 2\alpha_3 v.
\end{align}
The metric is given as
\begin{equation}
    g = 
    \begin{bmatrix}
        1 + h_u^2 & h_u h_v \\
        h_u h_v & 1 + h_v^2
    \end{bmatrix}.
\end{equation}
And its determinant as
\begin{align}
    |g| &= (1 + h_u^2)(1 + h_v^2) - (h_u h_v)^2 \\
    &= 1 + h_u^2 + h_v^2 + h_u^2 h_v^2 - h_u^2h_v^2 \\
    &= 1 + h_u^2 + h_v^2.
\end{align}
Finally, the inverse of $g$ can be computed as
\begin{align}
    g^{-1} = \frac{1}{|g|} 
    \begin{bmatrix}
        1 + h_v^2 & -h_u h_v \\
        -h_u h_v & 1 + h_u^2
    \end{bmatrix}.
\end{align}
Conveniently, at the center point $h_u(0, 0) = h_v(0, 0) = 0$ and thus
\begin{equation}
    g_{0, 0} = g^{-1}_{0, 0} = \mathbf{I}, \, |g_{0, 0}| = 1.
    \label{eq:metricidentity}
\end{equation}
To obtain nodes for the fitting of the quadratic polynomial, we project the points from a local neighborhood of $p$ onto the tangent plane.
The gradient of a function $X$ on the surface is given as
\begin{equation}
    \text{grad\,} X = \begin{bmatrix} \partial_u \g & \partial_v \g\end{bmatrix}g^{-1} \begin{bmatrix} \partial_u X \\ \partial_v X\end{bmatrix},
    \label{eq:gradient}
\end{equation}
where $\partial_u = \partial / \partial u$ is a shorthand for partial derivatives. Plugging \autoref{eq:metricidentity} into \autoref{eq:gradient}, we get
\begin{equation}
    \text{grad\,} X = \partial_u X \partial_u \g + \partial_v X \partial_v \g.
\end{equation}
$\partial_u \g$ and $\partial_v \g$ are exactly the basis vectors at point $p$. Thus, the coefficients of the resulting vectors are given by $\partial_u X$ and $\partial_v X$.
The function $X$ is given by function values at the points. 
To estimate the partial derivatives of $X$ at a point $p$, we locally fit a quadratic polynomial using the same approach as for fitting a quadratic polynomial to the surface and compute its partial derivatives. 
As for the fitting of the surface patch, we project the points in a local neighborhood to the tangent plane and use the function values as nodes for fitting the quadratic polynomial~\cite{Nealen2004}.

\paragraph{Discrete Divergence}
The divergence, including the metric components~\cite{o1983semi}, on the surface patch $\g$ is 
\begin{align}
    \text{div} V = \partial_u V_u + \partial_v V_v + V_u \partial_u \log \sqrt{|g|} + V_v \partial_v \log \sqrt{|g|},
    \label{eq:div}
\end{align}
where $|g|$ denotes the determinant of the metric.
At the origin, the metric of our surface patch is the identity and the derivatives of the metric at this point vanish. Hence, divergence is given by 
\begin{equation}
    \text{div\,}V = \partial_u V_u + \partial_v V_v.
    \label{eq:divergence}
\end{equation}

To compute the partial derivatives $\partial_u V_u,\partial_v V_v$ at $p_i$, we require the coefficients of the vector field at neighboring points $\{p_j \mid j \in \mathcal{N}_i\}$. However, different basis vectors are used at different points. Therefore, we need to map from the basis vectors at $p_j$ to those of $p_i$. While doing so, we account for metric distortion by $\g$. The following equation requires a bit more notation to distinguish between vectors at different points. We denote the coordinates of $p_j$ in the tangent space of $p_i$ as $(u_j, v_j)$, the metric of $\g$ at $p_j$ as $g_{u_j, v_j}$, the coefficients of a tangent vector at $p_j$ as $(\alpha_j^u, \alpha_j^v)$, and the basis vectors at $p_j$ as $\mathbf{e}_j^u, \mathbf{e}_j^v$. The coefficients of a vector at $p_j$ in $p_i$'s parameter domain are
\begin{equation}
    % \begin{bmatrix} (\alpha_j^u)_i \\ (\alpha_j^v)_i \end{bmatrix} = 
    g^{-1}_{u_j, v_j} \begin{bmatrix}\partial_u\g(u_j, v_j) \cdot \mathbf{e}_j^u & \partial_u\g(u_j, v_j) \cdot \mathbf{e}_j^v \\ \partial_v\g(u_j, v_j) \cdot \mathbf{e}_j^u & \partial_v\g(u_j, v_j) \cdot \mathbf{e}_j^v\end{bmatrix} \begin{bmatrix}\alpha_j^u \\ \alpha_j^v\end{bmatrix}.
    \label{eq:mapvector}
\end{equation}
\autoref{eq:divergence} and \autoref{eq:mapvector} are combined to form a sparse matrix $\mathbf{D} \in \mathbb{R}^{N \times 2N}$ representing divergence.

\section{Architectures}
We based our architectures on the designs proposed in DGCNN \cite{Wang2019}. A schematic overview is presented in \autoref{fig:architecture} and more details are provided in the following paragraphs.

\emph{Convolutions.} Each convolution, denoted as \textsc{Conv}($C_0, \ldots, C_L$), learns the function $h_{\mathbf{\Theta}}$ with an MLP that has $L$ layers. Each layer in the MLP consists of a linear layer with $C_i$ input- and $C_{i+1}$ output channels, batch normalization \cite{Ioffe2015}, and a non-linearity. For scalar features, the non-linearity is a leaky ReLU with slope $0.2$ and for vector features a ReLU. We denote MLPs applied per point as \textsc{MLP}($C_0, \ldots, C_L$).

\emph{Classification network.} The classification network has four convolution blocks: \textsc{Conv}(3, 64), \textsc{Conv}(64, 64), \textsc{Conv}(64, 128),  \textsc{Conv}(128, 256). Each scalar convolution is interspersed with connections to- and from the vector stream, which mirrors the number of parameters in its vector convolutions. The output of each scalar convolution is concatenated into a feature vector of $512$ features and transformed to $1024$ features using an MLP. We return a global embedding by taking both the maximum and mean of the features over all points. These are concatenated and fed to a task-specific head: \textsc{MLP}(2048, 512, 256, $C$), where $C$ is the number of classes in the dataset. This final MLP has dropout~\cite{dropout2014} set to $0.5$ in between the layers. During training, we optimize a smoothed cross-entropy loss.

\begin{figure}
    \centering
    \includegraphics[width=0.95\columnwidth]{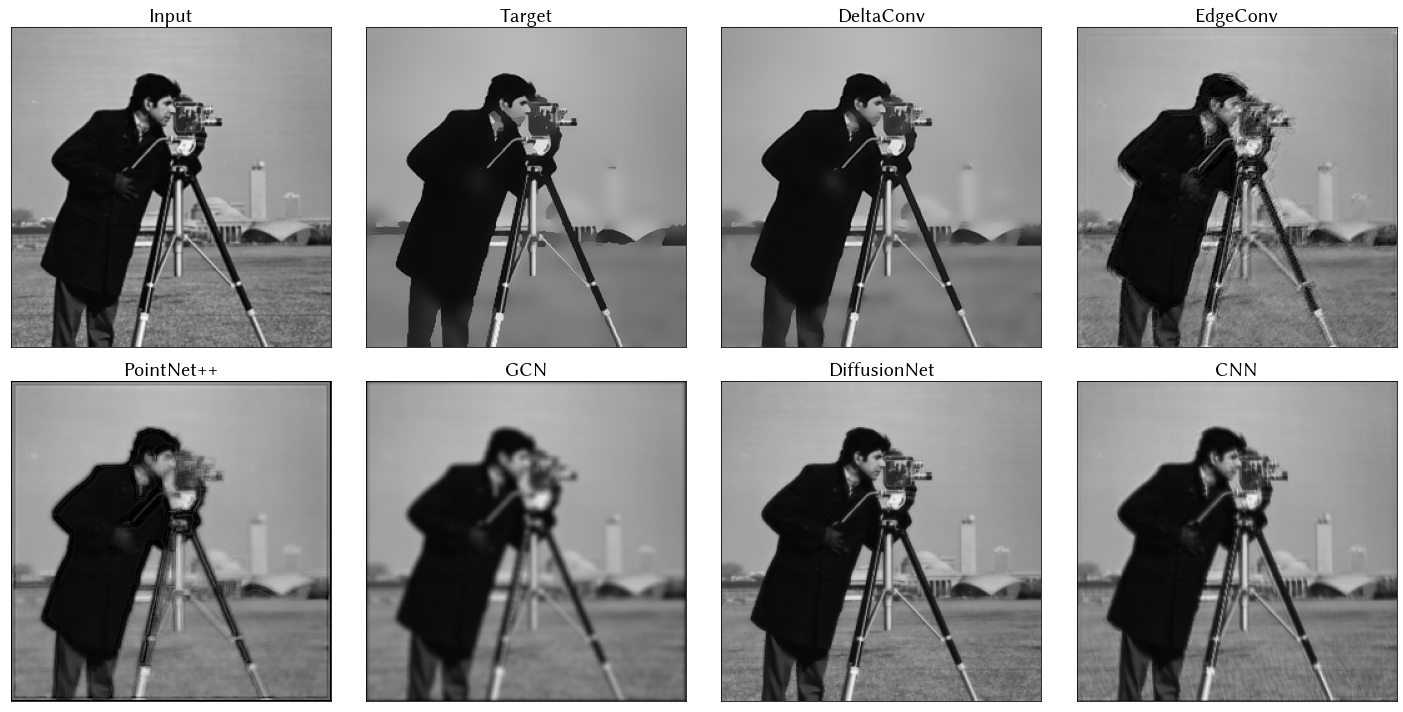}
    \caption{Comparison of different convolutions optimized to match the result of twenty anisotropic diffusion steps on sample image `camera'.}
    \label{fig:peronamalik_camera}
\end{figure}

\begin{figure}
    \centering
    \includegraphics[width=\columnwidth]{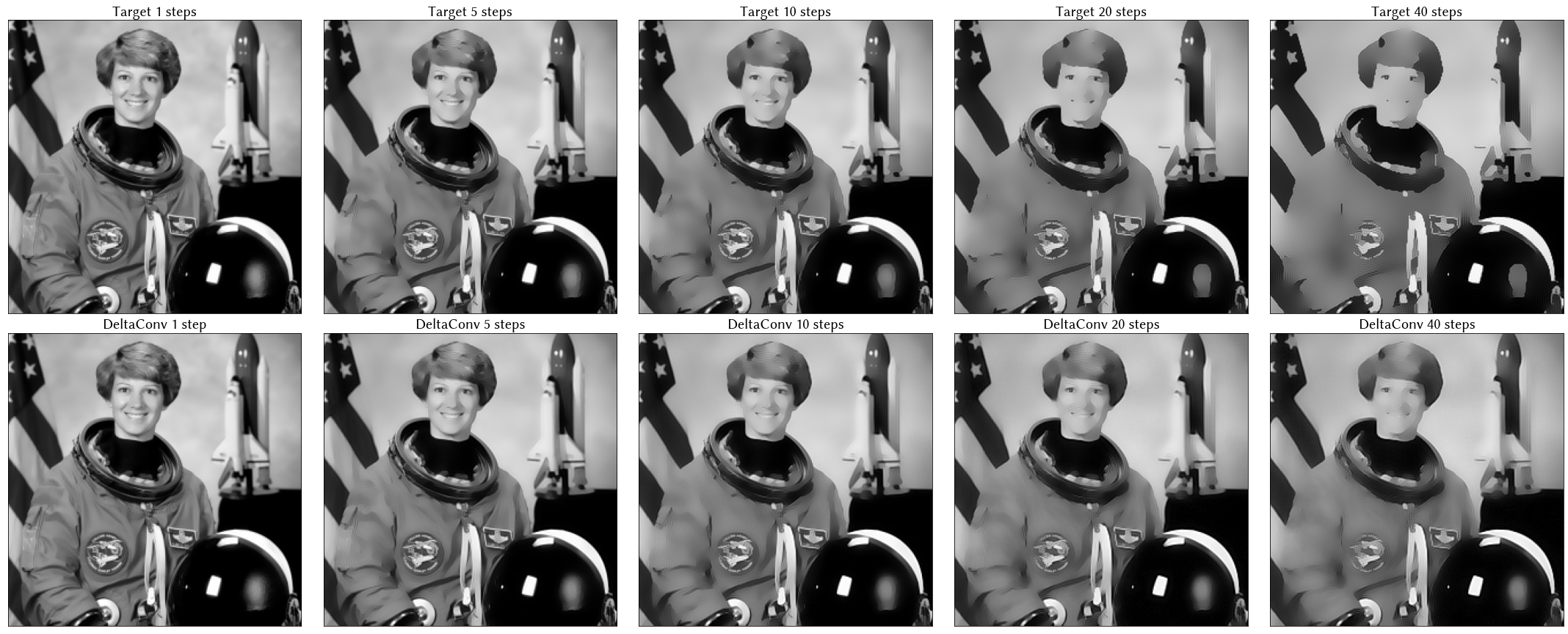}
    \caption{Comparison of a ResNet with DeltaConv optimized to match the result of varying anisotropic diffusion steps.}
    \label{fig:peronamalik_nsteps}
\end{figure}

\emph{Segmentation network.} The segmentation network uses three convolutions: \textsc{Conv}($C_{in}$, 64, 64), \textsc{Conv}(64, 128, 128),
\textsc{Conv}(128, 256, 256). Again, the scalar convolutions are interspersed with connections to- and from the vector stream. The output of each convolution is concatenated into a vector of $448$ features per point and transformed to $1024$ features with a global MLP. These features are pooled with maximum pooling. This embedding and an embedding of a one-hot encoding of the shape category is concatenated to the output of the convolutions at each point and fed to the task-specific head for segmentation: \textsc{MLP}(1536, 256, 256, 128, $C$). During training, we optimize a cross-entropy loss.

\emph{U-ResNet architecture} The U-ResNet architecture follows the design proposed in KPFCNN \cite{thomas2019KPConv} (Figure 9 of the supplementary material in \cite{thomas2019KPConv}). This network consists of an encoder that operates on four scales and a decoder that progressively upsamples the features to the original resolution. In each scale of the encoder, there are two ResNet blocks with a bottleneck. In KPFCNN, the first ResNet block uses strided convolutions, which we replace with pooling followed by a regular ResNet block. Each scale, we subsample to 1/4 points and increase the number of features by two. In the first layer, we use 64 features. We add two additional ResNet blocks with 128 output features after the decoder, as this was shown to be beneficial in CurveNet \cite{curvenet}. We do not use the other changes introduced by CurveNet, such as skip attention in the decoder or squeeze-excitation in the task-specific head. Each convolution block is replaced by a DeltaConv block, which maintains a vector stream in the first three scales and in the final two ResNet blocks. During pooling, scalar features are max-pooled and vector features are averaged with parallel transport to the coordinate system of the sampled point \cite{Wiersma2020}.

\begin{table}[t]
    \caption{
      Results on human part segmentation \cite{maron2017convolutional}.
    }
    \label{table:shapeseg}
    \begin{center}
        \resizebox{0.7\columnwidth}{!}{
        \begin{tabular}{lr}
            \toprule
            Method     & Accuracy \\
            \midrule
            PointNet++~\cite{Qi2017b} & 90.8 \\
            MDGCNN~\cite{Poulenard2018} & 88.6 \\
            DGCNN~\cite{Wang2019} & 89.7 \\
            SNGC~\cite{haim2019surface} & 91.0 \\
            HSN~\cite{Wiersma2020}  & 91.1 \\
            MeshWalker~\cite{Lahav2020MeshWalker} & \textbf{92.7}\\
            CGConv~\cite{yang2021continuous}  & 89.9 \\
            FC~\cite{mitchel2021field} & 92.5 \\
            DiffusionNet - xyz~\cite{Sharp2020DiffusionIA} & 90.6    \\
            DiffusionNet - hks~\cite{Sharp2020DiffusionIA} & 91.7     \\
            \midrule
            DeltaNet - xyz & 92.2 \\
            \bottomrule
        \end{tabular}
        }
    \end{center}
\end{table}
\begin{table*}
    \caption{Per-category breakdown of part segmentation results on ShapeNet part dataset. Metric is mIoU(\%) on points.}
    \label{table:part_segmentation_full}
    \begin{center}
        \resizebox{2.1\columnwidth}{!}{
            \begin{tabular}{l|c|ccccccccccccccccr}
                \toprule
                                & \textbf{Mean}      & aero          & bag           & cap           & car           & chair         & ear           & guitar        & knife         & lamp          & laptop        & motor         & mug           & pistol        & rocket & skate & table           \\
                                & \textbf{inst. mIoU} &               &               &               &               &               & phone         &               &               &               &               &               &               &               &        & board &                 \\
                \midrule
                \# shapes       &                    & 2690          & 76            & 55            & 898           & 3758          & 69            & 787           & 392           & 1547          & 451           & 202           & 184           & 283           & 66     & 152   & 5271          & \\
                \midrule
                PointNet++      & 85.1               & 82.4          & 79.0          & 87.7          & 77.3          & 90.8          & 71.8          & 91.0          & 85.9          & 83.7          & 95.3          & 71.6          & 94.1          & 81.3          & 58.7   & 76.4  & 82.6            \\
                PointCNN        & 86.1               & 84.1          & 86.5          & 86.0          & 80.8          & 90.6          & 79.7          & 92.3          & 88.4 & 85.3          & 96.1          & 77.2          & 95.3          & 84.2          & 64.2   & 80.0  & 83.0            \\
                DGCNN           & 85.2               & 84.0          & 83.4          & 86.7          & 77.8          & 90.6          & 74.7          & 91.2          & 87.5          & 82.8          & 95.7          & 66.3          & 94.9          & 81.1          & 63.5   & 74.5  & 82.6            \\
                KPConv deform   & 86.4               & 84.6          & 86.3          & 87.2          & 81.1          & 91.1          & 77.8          & \textbf{92.6} & 88.4 & 82.7          & 96.2          & \textbf{78.1} & 95.8
                                & 85.4               & \textbf{69.0} & \textbf{82.0} & 83.6                                                                                                                                                                                                             \\
                KPConv rigid    & 86.2               & 83.8          & 86.1          & 88.2          & \textbf{81.6} & 91.0          & 80.1          & 92.1          & 87.8          & 82.2          & 96.2          & 77.9          & 95.7          & \textbf{86.8} & 65.3   & 81.7  & 83.6            \\
                GDANet  & 86.5 & 84.2 & 88.0 & \textbf{90.6} & 80.2 & 90.7 & \textbf{82.0} & 91.9 & 88.5 & 82.7 & 96.1 & 75.8 & 95.7 & 83.9 & 62.9 & 83.1 & \textbf{84.4} \\
                PointTransformer & 86.6 & - & - & - & - & - & - & - & - & - & - & - & - & - & - & - & - \\
                PointVoxelTransformer & 86.5 & 85.1 & 82.8 & 88.3 & 81.5 & \textbf{92.2} & 72.5 & 91.0 & 88.9 & 85.6 & 95.4 & 76.2 & 94.7 & 84.2 & 65.0 & 75.3 & 81.7 \\
                CurveNet & 86.8 & 85.1 & 84.1 & 89.4 & 80.8 & 91.9 & 75.2 & 91.8 & 88.7 & \textbf{86.3} & 96.3 & 72.8 & 95.4 & 82.7 & 59.8 & 78.5 & 84.1 \\
                \midrule
                DeltaNet (ours) &  86.6 &	84.9 &	82.8 &	89.1 &	81.3 &	91.9 &	79.7 &	92.2 &	88.6 &	85.5 &	\textbf{96.7} &	77.2 &	\textbf{95.8} &	83.0 &	61.1 &	77.5 &	83.1            \\
                Delta-U-ResNet (ours) & \textbf{86.9} &	\textbf{85.3} & \textbf{88.1} &	88.6 &	81.4 &	91.8 &	78.4 & 92.0 & \textbf{89.3} & 85.6 & 96.1 & 76.4 & \textbf{95.9} & 82.7 & 65.0 & 76.6 &	84.1 \\
                \bottomrule
                \end{tabular}
        }
    \end{center}
    % \vskip -0.1in
\end{table*}

\section{Additional Results and Visualizations}

\subsection{Visualizations}
The anisotropic diffusion experiment was repeated for another input image with 20 anisotropic diffusion steps (\autoref{fig:peronamalik_camera}) and with varying anisotropic diffusion steps (\autoref{fig:peronamalik_nsteps}), showing that a DeltaConv network can approximate anisotropic diffusion for varying diffusion times.

\subsection{ShapeNet}
The per-category breakdown of ShapeNet is listed in \autoref{table:part_segmentation_full}.

\subsection{Human Shape Segmentation}
We trained a variant of the simple single-scale DeltaNet (eight layers with each 128 channels) to predict part annotations on the human body dataset proposed by \citet{maron2017convolutional}. This training set is composed of meshes from FAUST (100 shapes) \cite{faust2014}, SCAPE (71 shapes) \cite{scape2005}, Adobe Mixamo (41 shapes) \cite{adobe_2016}, and MIT (169 shapes) \cite{mit2008}. SHREC07 (18 shapes) is used for testing. Each dataset contains human bodies in different styles and poses, e.g., realistic, cartoony, dynamic. We convert the dataset into a point cloud dataset by uniformly sampling $8N$ points from the faces and downsampling these to $N$ points with FPS. We set $N=1024$, similar to the experiments in \citet{Wiersma2020}, $k=20$ and $\lambda=0.001$, similar to the other experiments. We normalize the area of the shape before sampling points and augment the input to the network with random rotations around the up-direction, a random scale between $0.8$ and $1.25$, and a random translation of 0.1 points. The network is optimized with Adam~\cite{adamkingma} for 50 epochs with an initial learning rate of $0.01$. The results are listed in \autoref{table:shapeseg}. This experiment shows DeltaConv's effectiveness on a deformable shape class and allows us to compare the results to those of other intrinsic (mesh) convolutions. This comparison has its limits, as most of the listed methods are trained on meshes instead of point clouds. Nonetheless, we find that DeltaConv is in line with state-of-the-art approaches, with only raw xyz coordinates as input.

\end{document}